\documentclass{article} %
\PassOptionsToPackage{table}{xcolor}

\PassOptionsToPackage{sort}{natbib}
\usepackage[final]{colm2025_conference}

\usepackage{microtype}
\usepackage{hyperref}
\usepackage{url}
\usepackage{booktabs}

\usepackage{lineno}

\definecolor{darkblue}{rgb}{0, 0, 0.5}
\hypersetup{colorlinks=true, citecolor=darkblue, linkcolor=darkblue, urlcolor=darkblue}

\usepackage{amssymb}
\usepackage{amsthm}

\usepackage{hyperref}
\usepackage{url}

\usepackage{inconsolata}

\usepackage{graphicx}
\usepackage{wrapfig}
\usepackage{subfigure}
\usepackage{booktabs} %

\usepackage{pgfplots}
\usepackage{pgfplotstable}
\pgfplotsset{compat=1.16} %

\usepackage{amsthm}

\usepackage{mathrsfs}
\usepackage{caption}
\usepackage{capt-of}
\usepackage{lipsum}
\usepackage{multibib}
\usepackage{adjustbox}
\newcites{app}{References}

\usepackage{tikz}

\definecolor{c1}{HTML}{765D97} 
\definecolor{c2}{HTML}{900C3F}
\definecolor{c3}{HTML}{fc6160}
\definecolor{myblue}{HTML}{E6F3FC} 
\definecolor{mygray}{HTML}{DBE2E9} 
\definecolor{mygreen}{HTML}{006400}

\hypersetup{
    colorlinks=true,
    linkcolor=black,
    urlcolor=c1,
    citecolor=c2,
}

\usepackage{amsmath,amsfonts,bm}

\newcommand{\figleft}{{\em (Left)}}

\newcommand{\figright}{{\em (Right)}}

\newcommand{\figtopleft}{{\em (Top Left)}}
\newcommand{\figbottomleft}{{\em (Bottom Left)}}
\newcommand{\figtopright}{{\em (Top Right)}}
\newcommand{\figbottomright}{{\em (Bottom Right)}}

\def\Figref#1{Figure~\ref{#1}}

\def\eqref#1{equation~\ref{#1}}

\def\1{\bm{1}}

\def\valpha{{\boldsymbol{\alpha}}}

\def\vgamma{{\boldsymbol{\gamma}}}

\def\MOmega{{\boldsymbol{\Omega}}}

\def\va{{\boldsymbol{a}}}

\def\vk{{\boldsymbol{k}}}

\def\vq{{\boldsymbol{q}}}

\def\vu{{\boldsymbol{u}}}
\def\vv{{\boldsymbol{v}}}

\def\vx{{\boldsymbol{x}}}
\def\vy{{\boldsymbol{y}}}

\DeclareMathAlphabet{\mathsfit}{\encodingdefault}{\sfdefault}{m}{sl}
\SetMathAlphabet{\mathsfit}{bold}{\encodingdefault}{\sfdefault}{bx}{n}

\newcommand{\R}{\mathbb{R}}

\newcommand{\head}[1]{\vspace{1.7mm}\noindent{{\textcolor{c1}{\bf #1.}}}}

\usepackage{tikz}

\definecolor{dark2orange}{rgb}{0.9, 0.4, 0.}
\definecolor{dark2purple}{rgb}{0.4, 0.4, 0.8}

\usepackage{mathtools}
\usepackage{multirow}

\usepackage{xspace}
\newcommand{\model}{Trellis\xspace}

\def\vrep{{\valpha}}
\def\rep{{\alpha}}
\newcommand{\nxt}{t}
\newcommand{\cur}{t-1}

\newcommand{\op}[1]{\operatorname{#1}}
\newcommand{\method}[1]{$\operatorname{#1}$}

\title{\textsc{\model}: 
Learning to Compress Key-Value Memory in \\ Attention Models}

\author{Mahdi Karami, Ali Behrouz,  Praneeth Kacham, Vahab Mirrokni 
\\
Google Research  \\
\texttt{\{mahdika,alibehrouz,pkacham,mirrokni\}@google.com} 
}

\begin{document}

\ifcolmsubmission
\linenumbers
\fi

\maketitle

\begin{abstract}
Transformers, while powerful, suffer from quadratic computational complexity and the ever-growing Key-Value (KV) cache of the attention mechanism.
This paper introduces \model,  a novel Transformer architecture with bounded memory that learns how to compress its key-value memory dynamically at test time. 
\model replaces the standard KV cache with a fixed-size memory and train a two-pass recurrent compression mechanism to store new keys and values into memory. 
To achieve this, it leverages an online gradient descent procedure with a forget gate, enabling the compressed memory to be updated recursively while learning to retain important contextual information from incoming tokens at test time.
Extensive experiments on language modeling, common-sense reasoning, recall-intensive tasks, and time series show that the proposed architecture outperforms strong baselines. 
Notably, its performance gains increase as the sequence length grows, highlighting its potential for long-context applications.

\end{abstract}

\section{Introduction}
Transformers~\citep{vaswani2018attention} has established itself as the \textit{de facto} architecture for sequence modeling in modern deep learning, achieving significant advances across diverse areas, including
language modeling~\citep{devlin2018bert,radford2018improving}, computer vision~\citep{dosovitskiy2020image, arnab2021vivit}, and graph learning and generation~\citep{yun2019graph,dwivedi2020generalization,karami2024higen}.
Their success stems from the attention mechanism, which allows models to dynamically attend to relevant parts of an input sequence while enabling parallel computation. This enables the capture of long-range dependencies and in-context learning.
However, the quadratic time and space complexity of attention with respect to sequence length restricts  its scalability in long sequence modeling.
Furthermore, the requirement for an \textit{unbounded cache} leads to inefficient memory management, particularly in resource-constrained environments. 

These limitations have driven the exploration of alternative architectures that aim to retain the representational power of Transformers while addressing their computational and memory complexity.
One key strategy involves \emph{sparsifying} the dense attention matrix through various techniques, including:  
blockwise attention~\citep{qiu2019blockwise, parmar2018image}; 
using strided or sliding window attention patterns~\citep{child2019generating, beltagy2020longformer, zaheer2020big}; 
or clustering/sorting tokens~\citep{kitaevreformer, roy2020efficient, tay2020sparse}.
Another approach involves \emph{low-rank approximations} of the self-attention matrix, leveraging the insight that it often exhibits low-rank properties~\citep{wang2020linformer}.
A different paradigm employs the kernel trick, replacing the Softmax operation with a dot product of feature maps, resulting in a family of linear attentions~\citep{katharopoulos2020transformers, choromanski2020rethinking, peng2021random}. 
While these methods substantially reduce computational costs, they may sacrifice expressiveness and performance, often requiring hybrid approaches that combine them with dense attention layers~\citep{mehta2022long, fu2023hungry}. 
Additionally, global convolutions~\citep{romero2021ckconv, li2022makes, poli2023hyena} and their input-dependent variants~\citep{karami2024orchid} have been explored as alternative sequence modeling techniques.

Recently, Recurrent Neural Network (RNN) architectures have re-emerged as promising attention-free solutions for sequence modeling.
These models leverage the parallelization capabilities of their linear recurrence,  building upon earlier linear time-invariant models~\citep{mehta2022long, wang2022pretraining, fu2023hungry} and extending them to more expressive input-dependent gated RNN designs with linear memory~\citep{orvieto2023LRU, gu2023mamba, de2024griffin, yang2024gatedattn} 
that demonstrate improved in-context learning while retaining computational advantages.
However, their effective memory is limited,  affecting their ability to efficiently compress and summarize information over very long contexts within their fixed-size hidden states.

\paragraph{Our approach and main contributions} 
In this paper, we present \model, a new Transformer architecture with bounded memory that learns how to compress its key-value memory at test time. To address the linearly-growing memory of global attention, \model{} replaces the KV-cache (i.e., the pair of $K$ and $V$ matrices) with a bounded memory with $m$ slots and train a meta in-context model to learn how to store new keys and values into memory. For better management of memory, \model{} uses a forgetting mechanism that learns how to selectively forget unnecessary past information in its compressed KV-cache.     

We perform an extensive set of experimental evaluations on various tasks including language modeling, commonsense reasoning, recall-intensive, needle-in-haystack, and time series data. We observe that \model{} outperforms state-of-the-art baselines, including Transformer++, and modern linear recurrent neural networks in downstream tasks. Furthermore, \model{} scales better than baselines when increasing the context length, showing promising results for long-context tasks.

\section{Related Work and Background}
For an input sequence $\mathcal{X} = [\vx_1, \dots, \vx_T]$, where $\vx_t \in \mathbb{R}^{d}$, the \emph{causal Softmax attention} mechanism generates output tokens $\vy_t \in \mathbb{R}^{d}$, by attending to past tokens (historic context):
\begin{align} \label{eqn:softmax}
\vy_t = \mathcal{V}_t ~ \op{Softmax}(\mathcal{K}_t^\top ~ \vq_t)  ~.
\end{align}
Here, the \textit{query}, \textit{key}, and \textit{value} vectors are computed by linear projections of the input: $\vq_t = \mathbf{W}_q~ \vx_t$, $\vk_t = \mathbf{W}_k ~\vx_t$, $\vv_t = \mathbf{W}_v~ \vx_t$, where $\mathbf{W}_q,$ $\mathbf{W}_k$, $\mathbf{W}_v$ $\in \mathbb{R}^{d \times d}$ are learnable weight matrices.
The key-value memory, represented by the caches $\mathcal{K}_t \in \mathbb{R}^{d \times t}$ and $\mathcal{V}_t \in \mathbb{R}^{d \times t}$, 
stacks the key and value vectors of each new token, leading to unbounded caches with linearly-growing size.
The retrieval of relevant information from this key-value cache can be rewritten as a weighted sum: 
\[
\vy_t = \mathcal{V}_t ~ \va_t, ~~
\text{where } \va_t = \op{Softmax}(\mathcal{K}_t^\top ~ \vq_t ) \in \mathbb{R}^{t}.
\]
Here, the vector $\va_t \in \mathbb{R}^{t} $
is the collection of the attention scores between $t$-th token and its historic context.
Hence, the attention in equation (\ref{eqn:softmax}) can be seen as a non-linear query from an unbounded memory.

The key-value cache size growth poses a significant memory bottleneck during inference, especially for long sequences. Additionally, each retrieval operation scales linearly with sequence length, resulting in an overall quadratic computational complexity $\mathcal{O}(T^2)$ for generating a full sequence of length $T$.

To address the computational and memory bottleneck of Softmax attention, various alternatives have been proposed~\citep{tay2022efficient}. 
A well-established approach involves employing the kernel trick to replace the Softmax with a dot product of feature maps, $\phi(\vq_t), ~ \phi(\vk_t)$ commonly known as Linear Attention \method{LA}~\citep{katharopoulos2020transformers}. 

To maintain bounded computational and memory requirements in attention mechanisms, an alternative strategy is to explicitly use a fixed-size key-value cache.  In this approach, the memory matrices $ \mathbf{K}$ and  $\mathbf{V}$ are constrained to a length $ m$ where $ m \ll T $.
A straightforward implementation of this strategy involves limiting the attention window to the most recent $ m$  tokens by maintaining a first-in-first-out (FIFO) queue, often referred to as Sliding Window Attention (SWA). While SWA achieves linear computational complexity, it suffers from a limited receptive field, restricting the model's ability to capture long-range dependencies and leading to a poor recall-memory trade-off~\citep{arora2024simple}.
Moreover, the observation, supported by many research works that key-value matrices in attention mechanisms often exhibit low-rank and sparse structures~\citep{wang2020linformer, chen2021scatterbrain, singhania2024loki} motivates the design of efficient sequence mixing layers that induce these properties.
Therefore, these layers aim to compress the context, storing the important information while discarding or forgetting redundancies, rather than naively truncating memory.

In light of this insight, the Attention-with-Bounded-Memory Control (ABC) mechanism~\citep{peng2021abc} introduces a method to compress and dynamically update a fixed-size memory. 
This is achieved using a pair of linear recurrences to update the key and value matrices:
\vspace{5pt}
\begin{align}\label{eq:abc-recurrence}
    {\mathbf{K}}_t &= {\mathbf{K}}_{t-1} + \valpha_t ~ \vk_t^\top \in \mathbb{R}^{m \times d}, ~ \nonumber\\ 
    {\mathbf{V}}_t &= {\mathbf{V}}_{t-1} + \valpha_t ~ \vv_t^\top \in \mathbb{R}^{m \times d},~ \nonumber\\
    \vy_t &= {\mathbf{V}}_t^\top \op{Softmax}({\mathbf{K}}_t \vq_t) \in \mathbb{R}^d
\end{align}
\vspace{5pt}

Here, $ \valpha_t := \op{Softmax}(\mathbf{W}_{\valpha} \vx_t) \in (0, 1)^m $ controls the update distribution across the memory slots. Each component, $\alpha_{t,j}$, can be interpreted as the writing intensity determining the $ t $-th token contribution to the $ j $-th memory slot.

The ABC update rule  can be decomposed into a cascade of two \method{LA}s and presented in the following two-pass process:
\vspace{5pt}
\begin{align} \label{eqn:ABC-2pass}
\{{\hat{\vy}}_t\}_{t=1}^T &= \operatorname{LA}(\{\vq_t, \vk_t, \valpha_t\}_{t=1}^T), ~~  {\hat{\vy}}_t \in \mathbb{R}^m, \nonumber \\
\{\vy_t\}_{t=1}^T &= \operatorname{LA}(\{f({\hat{\vy}}_t), {\valpha}_t, \vv_t\}_{t=1}^T), ~  \vy_t \in \mathbb{R}^d,  
\end{align}
\vspace{5pt}
where the intermediate activation function: $f(\cdot) = \op{Softmax}(\cdot)$. Building upon this foundation, Gated Slot Attention (GSA)~\citep{zhang2024gated} introduced a gated version of ABC, enhancing the two-pass process with a forget-gate mechanism introduced in GLA~\citep{yang2024gatedattn}.  This mechanism allows the model to forget irrelevant information, resulting in better memory management and improved performance.
Importantly, unlike many linear attention variants, both ABC and GSA do not rely on kernel approximations and retain the Softmax non-linearity.
Moreover, they can take advantage of the chunkwise matrix form~\citep{hua2022transformer, kacham2024polysketchformer, yang2024gatedattn}, enabling parallel hardware-efficient implementations on tensor cores. 

The simple linear additive (Hebbian-like) nature of the recurrence in ABC and GSA, however, limits the memory capacity and effective memory management in long-context retrieval tasks~\citep{schlag2021linear, behrouz2025itsconnected}. Specifically, the additive Hebbian update rule in ABC and GSA endlessly accumulates new tokens into the fixed memory space without an explicit mechanism to replace past value with new information in the memory, leading to an \emph{overcapacity regime}~\citep{schlag2021linear}. Moreover, their state-independent linear outer product modification of memory lacks dynamic interaction between the memory content and incoming keys, disabling it to selectively discard irrelevant or redundant information.

To address these limitations, this paper introduces a new two-pass \textit{non-linear recurrence}.
Leveraging techniques from meta-learning and test-time memorization~\citep{sun2024ttt, behrouz2024titans}, our approach is designed to dynamically compress new keys and values into the memory while minimizing information loss. 
For clarity, the subsequent explanation and its notations focus on key ($\vk_t$) cache  compression in the first pass, noting that the same principles apply to the value cache in the second pass.

\newpage
\section{Method}
We define the compression model as a \emph{regression layer} that projects a key token into a latent space. Specifically, for each token embedding ${\vk}_t$ and its corresponding latent representation ${\vrep}_t$,
the compression layer aims to reconstruct a target vector such that 
${\vrep}_t \approx \hat{\vrep}_t =  \phi(\mathbf{M}_t \vk_t)$.

To minimize the reconstruction error and the compression loss, we formulate the learning objective as an  $\ell_2$ optimization problem: 
\begin{align}
\mathcal{L}_t = \| \phi(\mathbf{M}_t \vk_t) -  {\vrep}_t \|^2, ~~   \mathbf{M}_t \in \mathbb{R}^{m \times d}, ~ \vk_t \in \mathbb{R}^{d}, ~ \vrep_t \in \mathbb{R}^{m}
\label{eq:decoding_loss}
\end{align}

We model the latent representation ${\vrep}_t$ using an encoder network, which is implemented as a linear projection of the input:
${\vrep}_t = \mathbf{W}_{\rep} \mathbf{x}_t $
with projection weight matrix $\mathbf{W}_{\rep} \in \mathbb{R}^{m \times d_x}$.
Our approach follows the framework of \emph{Fast Weight Programmers (FWPs)}~\citep{schmidhuber1992learning, schlag2021linear}, where the internal memory of the compression layer (\textit{a.k.a.} state in the context of RNNs), $\mathbf{M}_t$,  serves as ``fast weights", which are dynamically updated based on streaming input data. 
Hence, each sequence serves as a training dataset for the learning in this \emph{inner loop}.
To efficiently adapt to new tokens, we design an internal learning procedure that continuously updates $\mathbf{M}_t$, allowing it to store in-context information.

In this framework, the outer network, also referred to as the “slow” network, consists of the projection layer weights and the rest of the model parameters, jointly denoted as $\boldsymbol{\mathcal{W}}$. 
These parameters are trained in the outer loop, which follows standard deep neural network optimization process, minimizing the end-to-end loss averaged over the training dataset to learn generalizable patterns from the training set.  
This learning process subsequently enables
fast adaptation within the inner loop~\citep{schlag2021linear}.
Importantly, the weights of the “slow” network remain frozen during the internal state updates of the inner loop (involving the “fast” weights).  This overall procedure constitutes a bi-level optimization strategy~\citep{liu2022bome, chen2022gradient} commonly used in meta-learning (also referred to as learning to learn)~\citep{schmidhuber1992learning,thrun1998learning, bengio1990learning, andrychowicz2016learning}.

Given the sequential nature of the data, we approach this problem as an \emph{online optimization problem} and update the internal memory using one gradient descent step per token:
\begin{align} \label{eqn:OGD}
    \mathbf{M}_{t+1} = \mathbf{M}_{t} - \gamma_t \nabla_{\mathbf{M}} \mathcal{L}(\mathbf{M}_{t}, \vv_t, \vrep_t) 
\end{align}
This update rule generates the sequence of states $\{\mathbf{M}_t\}_{t=1}^T$, where each state $\mathbf{M}_t$ is a nonlinear recurrent function of the previous state and the input token. 
Consequently, the memory update follows a causal nonlinear recurrence, ensuring that information is 
continually integrated into the memory.

By the chain rule, the gradient of the loss with respect to $ \mathbf{M} $ can be computed:
\begin{align} \label{eqn:grad1}
 \nabla_{\mathbf{M}} \mathcal{L}_t 
 &= \mathbf{G}_t (\mathbf{M}_{\cur}, {\vrep}_{t} , {\vv_t}) 
 = 2 \, \left( \mathbf{J}_{\phi} \left( \phi(\mathbf{M}_{\cur} \, \vk_t) -  {\vrep}_t \right) \right)  \vk_t^\top
\end{align}
where $\mathbf{J}_{\phi}$ is the Jacobian of $ \phi(\cdot) $.
In practice, terms involving Jacobian products (such as 
$ \vu_t^\top := 
\mathbf{J}_{\phi} \left( \phi(\mathbf{M}_{\cur} \, \vk_t) -  {\vrep}_t \right)
$
in the expression above)  are computed efficiently using the {vector-Jacobian product} (\texttt{vjp}) method available in modern machine learning frameworks.
This avoids explicitly forming the full Jacobian matrix and leverages efficient automatic differentiation.

\subsection{State Decay}
While the proposed compression layer can address the quadratic time and space complexity of Transformers by learning how to effectively  compress key-value pairs $(k_t, v_t)$ into fixed-size memory states, 
these memories can still overfit to early tokens of the sequence or overflow as information accumulates. 
To mitigate these issues, we introduce $\ell_2$ regularization on the memory states—analogous to weight decay in standard neural network training—applied within the inner loop:
\begin{align}
\mathcal{L}_t = \| \phi(\mathbf{M}_t \, {\vv}_t) -  {\vrep}_t \|^2
+ \frac{\lambda_t}{2} \| \mathbf{M}_t \|^2, 
\label{eq:state_decay} 
\end{align}
This regularized objective results in a gradient descent recurrent update with \textit{state decay}:
\begin{align} \label{eqn:OGD_dacay}
    \mathbf{M}_{\nxt} = (1 - \lambda_t)\mathbf{M}_{\cur} - \gamma_t \nabla_{\mathbf{M}} \mathcal{L}(\mathbf{M}_{\cur}, \vv_t, \vrep_t) = \beta_t \mathbf{M}_{\cur} - \gamma_t \, \vu_t \, \vk_t^\top \in [0, ~1]
\end{align}
where, 
$\vu_t^\top := 
\mathbf{J}_{\phi} \left( \phi(\mathbf{M}_{\cur} \, \vk_t) -  {\vrep}_t \right)
$. 
This has been recently explored in the context of test-time memorization~\citep{wang2025testTimeRegression, karami2025lattice, behrouz2025itsconnected}. In this setting, the scalar  $ \beta_t = 1 - \lambda_t \in [0, ~1]$  acts as a forget gate, controlling the retention of prior memory.  
When $\beta_t \rightarrow 1$, it selectively updates the memory based on the interaction of state and input token without fading its magnitude, while $\beta_t \rightarrow 0$ erases the memory (possibly due to the change of context).
Such scalar gating mechanisms have gained renewed attention in recent RNN architectures as they provide a lightweight yet effective memory update~\citep{peng2021random,beck2024xlstm,sun2024you,yang2024gatedDeltaNet, behrouz2024titans}.

\subsection{Parallel and Hardware Efficient Implementation}

The non-linear nature of the update rule (\ref{eqn:OGD_dacay}) typically hinders straightforward parallelization. Several techniques have been proposed to address this limitation~\citep{gonzalez2024towards, lim2023parallelizing}.
\citet{sun2024ttt} introduces mini-batch gradient descent, where the sequence is divided into chunks, and the state at the beginning of each chunk is used to compute the gradients for all time steps within that chunk. In this approach, the gradients within each chunk are approximated as:
$\nabla_{\mathbf{M}} \mathcal{L}_t \approx \mathbf{G}_t (\mathbf{M}_{t'}, {\vrep}_{t} , {\vv_t}),
$
where $\mathbf{M}_{t'}$ represents the state at the beginning of the chunk (i.e., the final state from the preceding chunk and 
$t' \text{ = } t - \op{mod}(t, C)$ with $C$ denoting the chunk size. 

This strategy enables the parallel computation of a mini-batch of \emph{stale} gradients at the start of each chunk, thereby significantly enhancing scalability.
Using this approximation effectively linearizes the general non-linear recurrence (\eqref{eqn:OGD_dacay}) within each chunk, leading to the following recurrent update rule for the internal state and the memory readout of the compression layer:
\begin{align} \label{eqn:SDC_general}
\{\vy_t\}_{t=1}^T  &= \operatorname{compress}(\{\vq_t, \vk_t , \vrep_t\}_{t=1}^T)   
= 
\begin{cases}
    \mathbf{M}_{\nxt} = \beta_t \, \mathbf{M}_{\cur} 
    - 
    2 \gamma_t \, \left( \mathbf{J}_{\phi} \left( \phi(\mathbf{M}_{t'} \, \vk_t) -  {\vrep}_t \right) \right)  \vk_t^\top \\
    \vy_t = \mathbf{M}_{\nxt} {\vq}_{t}   
\end{cases}
\end{align}

This locally linear recurrence satisfies the associative property and can therefore be parallelized by parallel scan (a.k.a. prefix sum)~\citep{blelloch1990prefix}, or formulated into a parallel chunkwise form which has been shown to efficiently utilize the \texttt{matmul} units of the modern GPUs and can be more I/O efficient~\citep{hua2022transformer, kacham2024polysketchformer, yang2024gatedattn}.   
We adopt this approach and  derive a hardware-optimized chuck-wise form for the recurrence update with state decay in \eqref{eq:state_decay}.
For $b$-th block, covering time steps $t$ where $ bC+1 \le t < (b+1)C $, for $r$-th local step in the block, let's define the local cumulative product of decay from the start of the block as 
$\mu_r^b = \prod_{i=bC+1}^{bC+r} \beta_i$.
We also denote the segmented cumulative product  from step $i$ to $j$ within the $b$-th block ($1\le i\le j\le C$) as
$\omega_{j,i}^b = \frac{\mu_j^b}{\mu_i^b}$,
and the lower triangular matrix 
with entries $[\MOmega^b]_{j,i} = \omega_{j,i}^b~ \forall~ i \le j$ (and $0$ otherwise).
By unrolling the recurrence relation locally within the $b$-th block we obtain:
\begin{align}
\mathbf{M}_{r}^b = \beta_{r}^b\, \mathbf{M}_{r-1}^b - \gamma_{r}^b  \, {\vu_{r}^b}\, {\vk_{r}^b}^\top   
= \mu_r^b \mathbf{M}^{b-1} - \sum_{i=1}^{r} \gamma_{i}^b \, \MOmega_{r,i}   \,{\vu_{i}^b}\, {\vk_{i}^b}^\top   
\end{align}
Where $\mathbf{M}^{b-1}$ is the state at the end of the $(b-1)$-th block. Then $\mathbf{M}^{b}$ can be expressed in matrix form as:
\begin{align}
\mathbf{M}^b
= 
\mu^C \mathbf{M}^{b-1} -  
{\mathbf{U}^b} \, \op{Diag}({\vgamma} \odot \MOmega_{B:}) \, {\mathbf{K}^b}^\top
\end{align}
where, $\odot$ is the element-wise product. Also, unrolling the readout step and formulating it in matrix form for the entire block yields:  
\begin{align}
\mathbf{Y}^b
= \op{Diag} (\boldsymbol{\mu}^b) \, {\mathbf{M}^{b}} {\mathbf{Q}^b}
-
{\mathbf{U}^b} \left( {\mathbf{K}^{b}}^\top {\mathbf{Q}^b}  \odot \op{Diag}(\vgamma) \odot \MOmega  \right)  
\end{align}
where $\boldsymbol{\mu}^b = [\mu_1^b, \dots, \mu_C^b]$, and 
$\mathbf{K}^b,~\mathbf{Q}^b \in \R^{d \times C }$ and $\mathbf{U}^b,~\mathbf{Y}^b \in \R^{m \times C}$ are matrices collecting the corresponding vectors within the block.
This matrix form, which computes the states only at the end of each block, extends the State Space Duality formulation introduced in Mamba2~\citep{mamba2} 
 and enables efficient use of the \texttt{matmul} operations on GPUs. 

Consequently, applying the proposed recurrent compression model to key and value caches in a two-pass process yields the following operations:
\begin{align}
\{\hat{\vy}_t\}_{t=1}^T &= \operatorname{compress}(\{\vq_t, \vk_t, \valpha_t\}_{t=1}^T), 
\quad 
\{ \hat{\vy}_t, {\valpha}_t, \} \in \mathbb{R}^m 
\label{eq:comp-1st-pass}
\\
\{\vy_t\}_{t=1}^T &= \operatorname{compress}(\{f(\hat{\vy}_t), \vv_t, \valpha_t\}_{t=1}^T), 
\quad  \vy_t  \in \mathbb{R}^d,  
\label{eq:comp-2nd-pass}
\end{align}

We explored alternative choices for the intermediate activation function in our architecture and found that \textit{normalized SiLU} defined as  
$f(\vx) ~\text{=}~ \frac{\op{SiLU}(\vx)}{\|\op{SiLU}(\vx)\|}$
outperforms the commonly used $\op{Softmax}$, also known as the normalized exponential function. 
This improvement can be explained by the fact that, in our architecture—where the cache is densely compressed into a limited number of memory slots ($m \ll T$)—a normalization function with \textit{less spikiness than $\operatorname{Softmax}$} is more effective for retrieval.
It is also worth noting the difference in input ordering between the second pass of our architecture and that of ABC and GSA.  In \model{}, $\vrep_t$ acts as a shared target vector for both compression layers. Consequently, we use  $\vy_t = \phi(\mathbf{M}_{t}^{\top} {\vq}_{t})$  as the readout operation of the value compression layer to ensure correct output dimensions.
As a result, in the first pass (\eqref{eq:comp-1st-pass}), \model compresses the new key embedding, $\vk_t$, into its memory, generating an intermediate representation and in the second pass (\eqref{eq:comp-2nd-pass}), it compress the new value embedding and finally outputs  $\vy_t$.

\begin{figure*}[t]
    \centering
    \raisebox{0.5\height}{
    \includegraphics[trim= 25 50 140 0,clip,width=.7\linewidth]{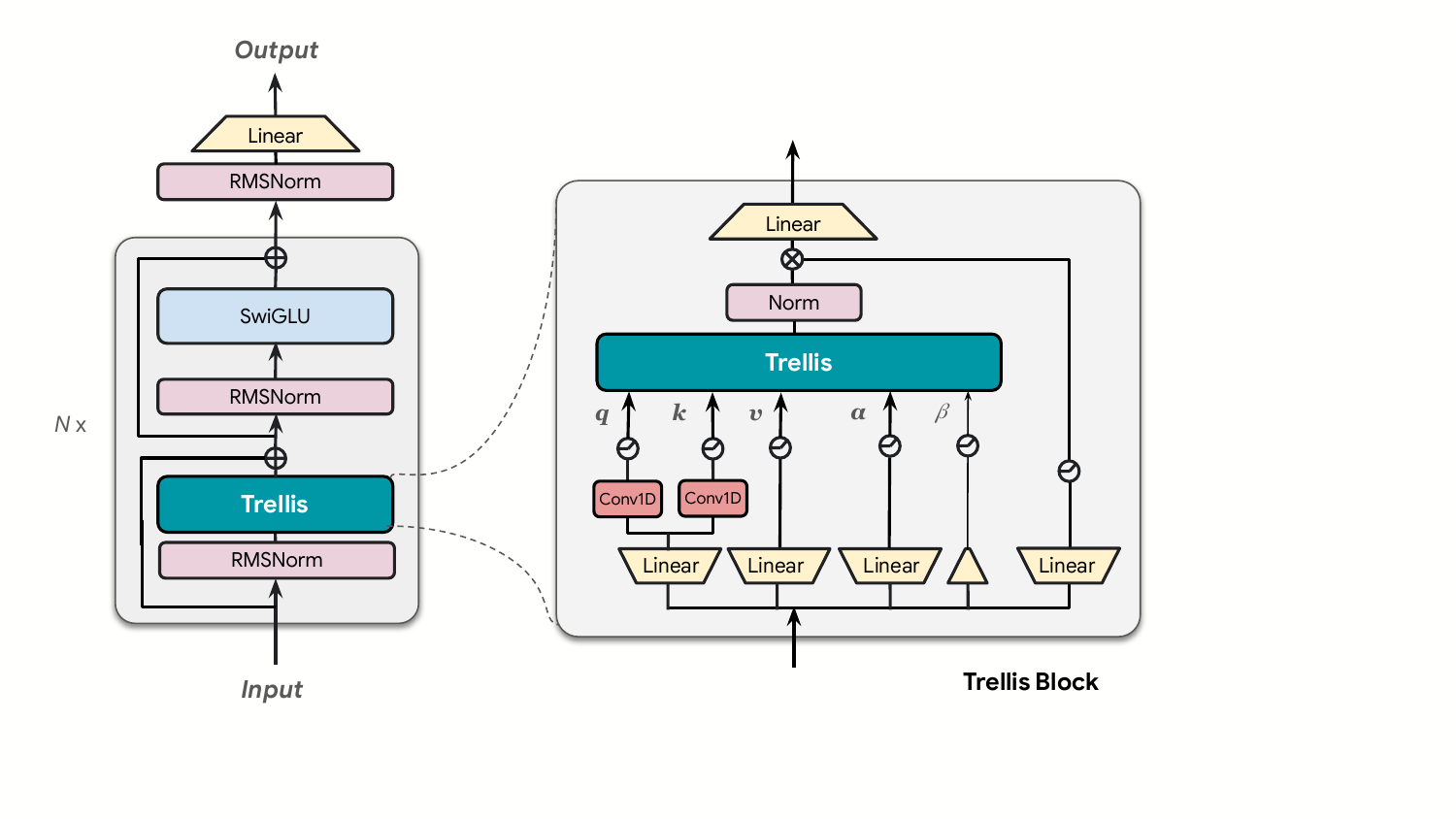}
    }
\vskip -70pt
  \caption{\captionsize
    \figleft~Block diagram of the language model.
    \figright~The \model block. Each sequence mixing block is composed of a short $\operatorname{Conv1D}$ for $\{q, k \}$ and the \model is followed by a post normalization and the $\operatorname{GeLU}$ post-gate.
    }
  \label{fig:model}
\end{figure*}

The overall \model{} architecture used for language models is illustrated in \Figref{fig:model}.

\subsection{Related Works and Discussion}

\textbf{Linear State Space Models (SSMs) and Recurrent Neural Networks (RNNs)} have recently received renewed interest as an efficient paradigm for sequence modeling. Offering sub-quadratic scaling during training and constant-time recurrence at inference, they have proven particularly effective for modeling long-range dependencies~\citep{gu2020hippo}. 
The recurrence in the linear time invariant SSMs can 
be reformulated as a global convolution, enabling efficient computational implementations~\citep{gu2021combining, mehta2022long}. 
Furthermore, the linearity in input-dependent SSMs~\citep{gu2023mamba, mamba2} and modern gated RNNs~\citep{orvieto2023LRU, de2024griffin, beck2024xlstm,karami2025MSSSM} take advantage of
parallelization through techniques like associative scan~\citep{blelloch1990prefix, smith2023simplified, de2024griffin}, 
or chunkwise parallel forms~\citep{hua2022transformer, yang2024gatedattn, yang2024parallelizing, behrouz2024titans, behrouz2025atlas}.
However, a potential limitation arises from state-independent updates. The additive (gated) linear modification of the memory,
doesn't consider the interaction between the memory content and incoming keys, potentially limiting the model's ability to efficiently compress and summarize information within its finite memory state.

\textbf{Fast Weight Programmers (FWPs)} represent a class of architectures where the parameters of one network, often termed the ``fast" network, are dynamically generated or modified by another ``slow" network~\citep{schmidhuber1992learning, schlag2021linear}. This concept, also referred to as input-dependent parameterization~\citep{karami2019invertible, gu2023mamba,karami2024orchid}, enables the model to adapt to the specific characteristics of its input, potentially capturing more complex contextual patterns. 
In our framework, the compression layer's memory state acts as ``fast weights" that quickly learn in-context information, while  the ``slow" network parameters learns generalizable patterns across the training set.
Therefore, we effectively adopt a meta-learning~\citep{schmidhuber1992learning,thrun1998learning, bengio1990learning, andrychowicz2016learning} style approach by deriving an internal online learning procedure that continuously minimizes a reconstruction loss.

\textbf{Online Gradient Descent} is widely adopted for non-stationary sequential data in classical adaptive filtering such as the Least Mean Squares (LMS) algorithm~\citep{haykin2002adaptive}. LMS-also known as the \emph{Delta Rule}~\citep{widrow1988adaptive, schlag2021linear}-typically minimizes the instantaneous squared error through a simple a linear update rule.
Convex optimization-based methods have also been explored for identifying SSMs (a.k.a. linear dynamical systems); for example,  ~\citet{karami2017multi} formulates the problem as a multi-view matrix factorization and proposes a global optimizer. 

In contrast, our work offers a more general non-linear update rule incorporating a forgetting gate. This update rule is applied to both key and value compression within a cascading, two-pass structure. This design is particularly suitable for long-context modeling, where efficient and expressive memory compression is critical.

\section{Experiments}\label{sec:expriments}

\paragraph{Overview:}
In this section, following recent studies~\citep{lim2023parallelizing, zhang2024gated}, we evaluate the performance of \model{} in various downstream tasks, including language modeling, common-sense reasoning, needle in haystack, and time series forecasting. We use four scales of the proposed \model{} with (1) 125M,  (2) 350M, (3) 780M, and (4) 1B parameters. For each scale, we train three different versions, each of which is solely trained on either Pile~\citep{gao2020pile}, C4~\citep{raffel2020exploring}, or Books datasets (a subset of the Pile). The main reason for our choice of datasets and training different version of \model{} are two folds: First, the dataset used for training and its characteristics can significantly affect the performance of the model. Therefore, while a dataset can be more useful for a model, it might not be the best choice for others. Accordingly, to avoid cherry-picking of the dataset, and to emphasize the generalizebility and the power of \model, we use three different versions, each of which is trained on a commonly-used dataset in the literature.  Second, we use each dataset to highlight the strength of \model{} in one aspects. 
That is, we use Pile (resp. Books) dataset to highlight the strength of \model{} in short context (resp. long context).
\paragraph{Baseline Models: }  
We compare our method against the \method{Transformer}++ architecture proposed in~\cite{touvron2023llama} and the following sub-quadratic models: 
\method{Linear-Attention~(LA)}~\citep{katharopoulos2020transformers}, 
\method{TTT}~\citep{sun2024ttt},
\method{GSA}~\citep{zhang2024gated},
\method{DeltaNet}~\citep{yang2024parallelizing},
\method{Gated-DeltaNet}~\citep{yang2024gatedDeltaNet},
\method{Mamba2}~\cite{mamba2}.
All RNN models follow the \method{Mamba} architecture where the sequence models follows by a normalization and gating before output linear projection.

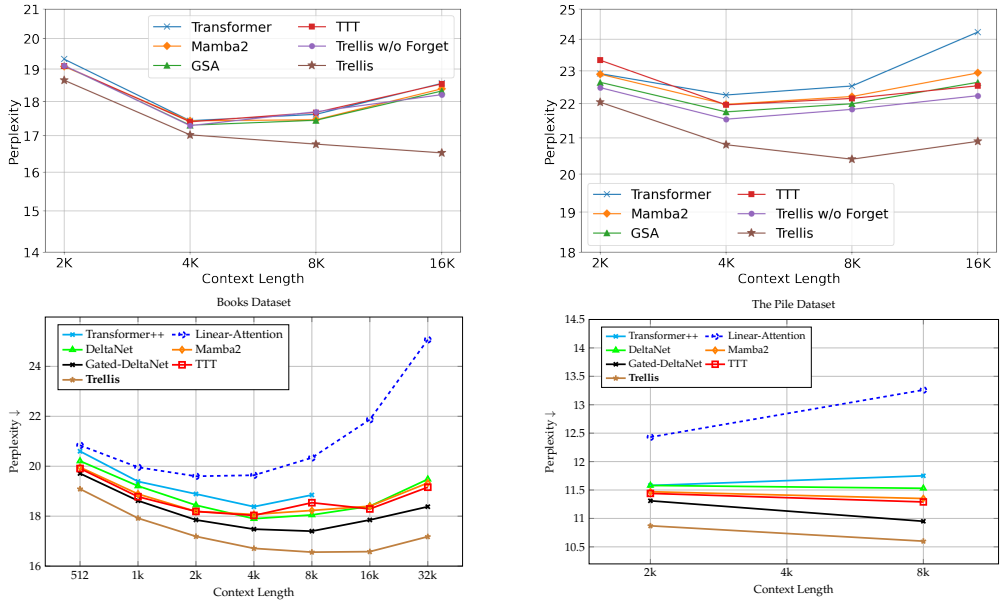
\begin{figure}
\centering
\begin{minipage}{.49\textwidth}
  \centering
    \includegraphics[width=0.9\linewidth]{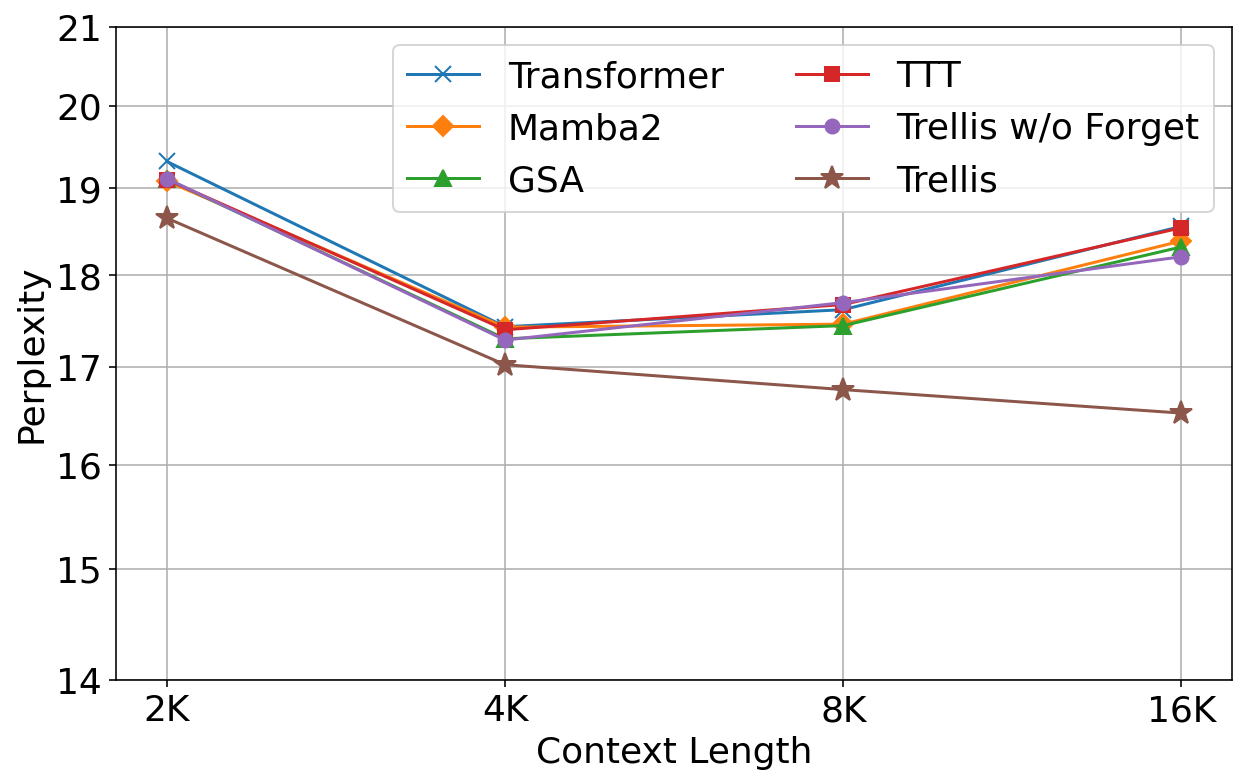}
\end{minipage}~\hfill{}~
\begin{minipage}{.49\textwidth}
  \centering
    \includegraphics[width=0.9\linewidth]{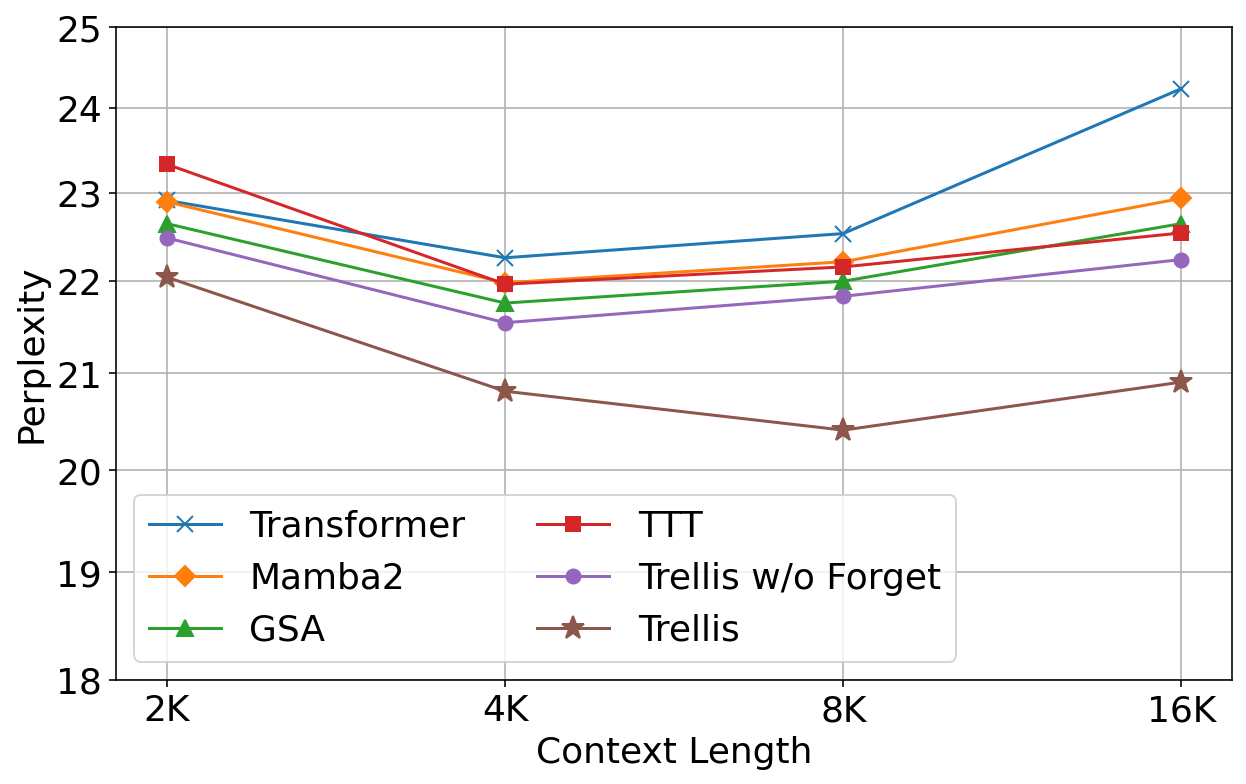}
\end{minipage}
\begin{minipage}{.49\textwidth}
  \centering
  \begin{adjustbox}{width=.9\linewidth}
      \pgfplotsset{
    compat=1.16, %
    width=14cm, %
    height=9cm, %
    every axis plot/.append style={
        line width=1.5pt, %
        solid,            %
        mark size=2.5pt,  %
    }
}

\begin{tikzpicture}
    \begin{axis}[
        title={Books Dataset},
        xlabel={Context Length},
        ylabel={Perplexity  $\downarrow$},
        xmode=log, %
        log base x=2, %
        xtick={512, 1024, 2048, 4096, 8192, 16384, 32768},
        xticklabels={512, 1k, 2k, 4k, 8k, 16k, 32k},
        grid=major, %
        ymin=16, %
        legend columns=2,
        legend pos=north west, %
        legend cell align={left},
        grid=major, 
        ymode=linear,           %
    ]

    \pgfplotstableread{ 
    Context Transformer LinearAtt DeltaNet Mamba2 GatedDelta TTT OurModel
    512     20.60     20.84     20.21    19.96   19.71      19.91 19.09
    1024    19.39     19.96     19.21    18.89   18.62      18.78 17.92
    2048    18.89     19.60     18.44    18.19   17.85      18.19 17.19
    4096    18.38     19.64     17.90    18.07   17.48      18.03 16.71
    8192    18.85     20.34     18.05    18.23   17.40      18.54 16.56
    16384     nan     21.87     18.40    18.40   17.85      18.29 16.58
    32768   nan       25.07     19.48    19.33   18.38      19.16 17.18
    }\datatable

    \addplot+[color=cyan, mark=x] table [x=Context, y=Transformer] {\datatable}; %
    \addlegendentry{\method{Transformer}++}

    \addplot+[color=blue, mark=o, dashed] table [x=Context, y=LinearAtt] {\datatable}; %
    \addlegendentry{\method{Linear-Attention}}

    \addplot+[color=green, mark=triangle, solid] table [x=Context, y=DeltaNet] {\datatable};
    \addlegendentry{\method{DeltaNet}}

    \addplot+[color=orange, mark=diamond] table [x=Context, y=Mamba2] {\datatable};
    \addlegendentry{\method{Mamba2}}

    \addplot+[color=black, mark=x, solid] table [x=Context, y=GatedDelta] {\datatable};
    \addlegendentry{\method{Gated-DeltaNet}}

    \addplot+[color=red, mark=square, solid] table [x=Context, y=TTT] {\datatable};
    \addlegendentry{\method{TTT}}

    \addplot+[color=brown, mark=star, solid] %
        table [x=Context, y=OurModel] {\datatable};
    \addlegendentry{\textbf{\model}}

    \end{axis}
\end{tikzpicture}
  \end{adjustbox}
\end{minipage}~\hfill{}~
\begin{minipage}{.49\textwidth}
  \centering
  \begin{adjustbox}{width=.9\linewidth}
    \pgfplotsset{
    compat=1.16, %
    width=14cm, %
    height=9cm, %
    every axis plot/.append style={
        line width=1.5pt, %
        solid,            %
        mark size=2.5pt,  %
    }
}

\pgfplotstableread{
Context Transformer LinearAtt DeltaNet Mamba2 GatedDelta TTT OurModel
2048    11.58     12.43     11.58    11.47   11.31      11.44 10.87
8192    11.75     13.26     11.53    11.35   10.95      11.29 10.60
}\datatablepile %

\begin{tikzpicture}
    \begin{axis}[
        title={The Pile Dataset},
        xlabel={Context Length},
        ylabel={Perplexity $\downarrow$},
        xmode=log, %
        log base x=2, %
        xtick={512, 1024, 2048, 4096, 8192, 16384},
        xticklabels={512, 1k, 2k, 4k, 8k, 16k},
        xmin=1500,
        xmax=12000,
        ymin=10.2, %
        ymax=14.5, %
        legend columns=2,
        legend pos=north west, %
        legend cell align={left},
        legend style={font=\small},
        grid=major,             %
        ymode=linear,           %
    ]

    \addplot+[color=cyan, mark=x] table [x=Context, y=Transformer] {\datatablepile}; %
    \addlegendentry{\method{Transformer}++}

    \addplot+[color=blue, mark=o, dashed] table [x=Context, y=LinearAtt] {\datatablepile}; %
    \addlegendentry{\method{Linear-Attention}}
    
    \addplot+[color=green, mark=triangle, solid] table [x=Context, y=DeltaNet] {\datatablepile};
    \addlegendentry{\method{DeltaNet}}

    \addplot+[color=orange, mark=diamond] table [x=Context, y=Mamba2] {\datatablepile};
    \addlegendentry{\method{Mamba2}}

    \addplot+[color=black, mark=x, solid] table [x=Context, y=GatedDelta] {\datatablepile};
    \addlegendentry{\method{Gated-DeltaNet}}

    \addplot+[color=red, mark=square, solid] table [x=Context, y=TTT] {\datatablepile};
    \addlegendentry{\method{TTT}}

    \addplot+[color=brown, mark=star, solid] %
        table [x=Context, y=OurModel] {\datatablepile};
    \addlegendentry{\textbf{\model}}

    \end{axis}
\end{tikzpicture}
  \end{adjustbox}
\end{minipage}
  \captionof{figure}{The effect of context length on model's perplexity. Subplots show:
  \figtopleft~ C4 dataset with  350M parameters; 
  \figtopright~ C4 dataset with  780M parameters;
  \figbottomleft~ Books dataset with  125M parameters; 
  \figbottomright~ The Pile dataset with  125M parameters.
  Training Transformers from scratch on very long sequence contexts (e.g., $T \in \{16k, 32k\}$) can yield poor perplexity, hence the standard practice for such contexts is typically to finetune a Transformer pre-trained on shorter sequences~\citep{touvron2023llama}. Here, for the Transformer baseline trained from scratch in these results, performance is only reported up to a context length of $T=8k$.
  }
    \label{fig:contextlength}
\end{figure}

To investigate the effect of context length on model performance, in this part, we evaluated models trained with various context lengths, comparing \model{} (both with and without its forget gate) against the baseline models. For the Books dataset, models were trained using sequence lengths of
 $\{512, 1024, 2048, 4096, 8192, 16384, 32748\}$. 
We also trained the models on the C4 dataset with context lengths in $\{2048, 4096, 8192, 16384\}$, and on the Pile dataset using a subset of context lengths $ T \in \{2048, 8192\}$. 
The results presented in \autoref{fig:contextlength} show that \model{} achieves the lowest perplexity compared to all baselines across all tested context lengths. Interestingly, \model{} shows greater performance gains compared to other linear RNNs as the sequence length increases, 
demonstrating the potential of our approach for tasks requiring long-context reasoning. Furthermore, comparing  against \model{} w/o forgetting highlights the importance of this component in the overall performance of \model{}.

\paragraph{Language Modeling and Common-sense Reasoning}
Following recent studies on sequence modeling~\citep{behrouz2024titans, yang2024gatedDeltaNet}, in this section, we compare the performance of \model{} with modern linear recurrent neural networks and Transformer on language modeling and common-sense reasoning tasks. The results are reported in \autoref{tab:lm_results}. \model{} achieves outstanding performance across all scales and outperform all linear recurrent models and Transformer++.

\begin{table*}[t!]
\centering
\caption{
Performance of \model{} and baselines on language modeling and common-sense reasoning tasks. 
}\label{tab:lm_results}
\centering
\resizebox{0.85\linewidth}{!}{
\centering
\begin{tabular}{l|c|c c c c c c c c c}
\toprule
\textbf{Model}  &  \textbf{LMB.} &  \textbf{LMB.} & \textbf{PIQA} &    \textbf{Hella.} & \textbf{Wino.} & \textbf{ARC-e} &  \textbf{ARC-c} &  \textbf{SIQA}  &  \textbf{Avg.} \\
 & ppl $\downarrow$  &  acc $\uparrow$  & acc $\uparrow$ &   acc $\uparrow$  & acc $\uparrow$  & acc $\uparrow$ & acc $\uparrow$ &  acc $\uparrow$  &   $\uparrow$  \\
\midrule
\midrule
\multicolumn{10}{c}{790M params / 30B tokens} \\
\midrule
 \method{Transformer}++ & 25.89 & 33.41 & 64.75 & 41.98 & 51.33 & 59.06 &  31.85 & 40.26 & 46.09 \\
 \method{Mamba2} & 28.91 & 32.72 & 64.98 & 42.60 & 50.01 & 61.99 & 30.24 & 41.07 & 46.23\\
 \method{TTT} & 27.05 & 33.18 & 65.03 & 43.17 & 49.93 & 62.16 & 32.13 & 41.35 & 46.71 \\
  \method{Gated-DeltaNet} & 21.40 & 34.83 & 65.79 & 43.66 &  50.45  & 64.02 & 32.24 & 41.68 & 47.52 \\
  \midrule
\method{\model} & 20.28 & 35.44 & 67.51 & 44.29 & 51.08 &  65.12 & 33.17 & 42.04 & 48.38 \\
\bottomrule
\end{tabular}
}
\end{table*}

\begin{figure*}
\centering
\begin{minipage}[t]{.32\textwidth}
  \centering
        \centering
\footnotesize
    \captionof{table}{ \label{tab:ablation}
    \captionsize
    Ablations on improving from linear  \method{DeltaNet}~\citep{yang2024parallelizing} and also \method{TTT}~\citep{sun2024ttt}. All models have 125M parameters trained on The Pile dataset. 
    }
\setlength{\tabcolsep}{4pt}
\renewcommand{\arraystretch}{1.3}
\centering
\resizebox{0.99 \linewidth}{!}{
    \begin{tabular}{lc}
        \toprule
        \textbf{Configuration} & ppl $\downarrow$ 
        \\
        \midrule
        \method{DeltaNet}
        & 11.58  \\    
        \method{TTT}
        & 11.44  \\        
        \midrule
        \method{\model} &  10.87   \\
        \quad \method{w/o~forget~gate} & 11.28   \\
        \quad \method{with~f=\texttt{L2-SiLU}} & 10.98   \\
        \quad \method{with~f=\texttt{Softmax}} & 11.29   \\
        \quad \method{Linear~f=\texttt{Softmax}} & 12.71   \\
        \quad \method{Linear~f=\texttt{LN-SiLU}} & 11.65   \\
        \quad \method{m=32} & 11.14   \\
        \quad \method{m=128} & 10.87   \\
        \quad \method{b=1} &  10.75   \\
        \bottomrule
    \end{tabular}
}

\end{minipage}~\hfill{}~
\begin{minipage}[t]{.66\textwidth}
  \centering
      \centering
    \captionof{table}{Performance of \model{} and baselines with 1B parameters on S-NIAH task from RULER benchmark. The best results with highest accuracy are highlighted.
    }
    \label{tab:haystack}
    \resizebox{0.99 \linewidth}{!}{
    \begin{tabular}{l c c c c c c c c c c}
    \toprule
    \multirow{2}{*}{Model} & \multicolumn{3}{c}{\textbf{S-NIAH-PK}} & \multicolumn{3}{c}{\textbf{S-NIAH-N}} & \multicolumn{3}{c}{\textbf{S-NIAH-W}} & \multirow{2}{*}{\textbf{Average}} \\
    \cmidrule(lr){2-4} \cmidrule(lr){5-7} \cmidrule(lr){8-10}
    &  2K & 4K & 8K &  2K & 4K & 8K  & 1K & 2K & 4K  \\
    \midrule
    \midrule
       TTT  & 98.4 & \cellcolor{myblue}98.8 & 98.0 & 60.2 & 36.6 &  10.2 & 85.8  & 78.8 & 28.0 & 66.1 \\
       Mamba2 & 98.6 & 61.4 & 31.0 & 98.4 & 55.8 & 14.2 & 62.2  & 42.2 & 4.2 & 52.0 \\
       DeltaNet & 96.8 & \cellcolor{myblue}98.8 & \cellcolor{myblue}98.6 & 47.2 & 15.4 & 12.8 &  85.2 & 46.2 & 20.0 & 57.9 \\
       Gated DeltaNet & 89.8 & 91.4 & 90.0 & 99.2 & 91.8 & 26.4 & \cellcolor{myblue}86.4 & 82.6 & 24.4 & 75.8\\
       \midrule
       \model & \cellcolor{myblue}99.2  & 95.2 & 97.8 & \cellcolor{myblue}99.4 & \cellcolor{myblue}94.2 & \cellcolor{myblue} 34.6 & \cellcolor{myblue}86.4 & \cellcolor{myblue}82.8 & \cellcolor{myblue}28.4 & \cellcolor{myblue} 79.8 \\
    \toprule
    \end{tabular}
    }

    \vskip 5pt
  \includegraphics[width=0.8\linewidth]{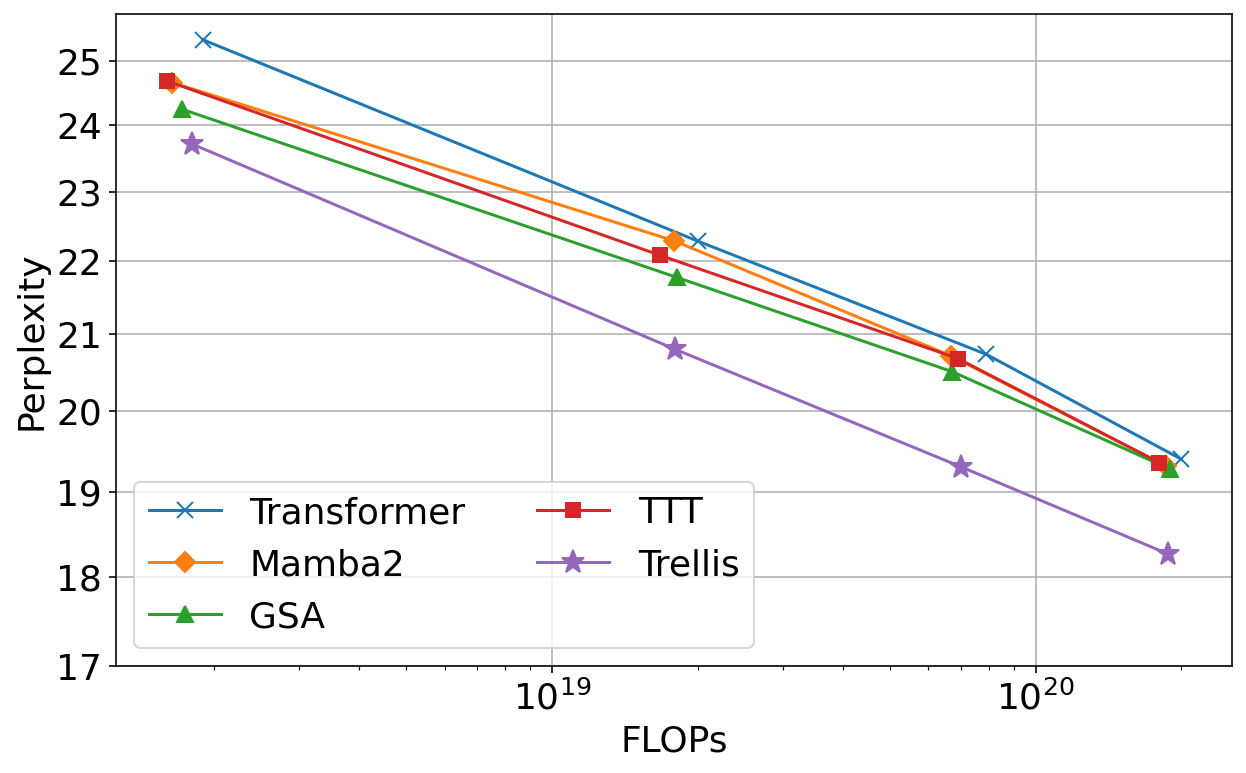}
    \vspace{5pt}
  \captionof{figure}{\captionsize
    Scaling pattern of models w.r.t. Perplexity vs. FLOPs.}
  \label{fig:scaling}
\end{minipage}
\end{figure*}

To see the scaling law in \model{} and compare it with baselines, we report the perplexity for different model sizes in \autoref{fig:scaling}. \model{} shows a consistent trend and achieves better perplexity compared to baselines with a fixed budget of FLOPs. This pattern shows that in the trade-off of efficiency and effectiveness, \model{} achieves Pareto frontier results. 

\paragraph{Needle in Haystack Tasks}
In this section, we follow recent studies on sequence modeling and evaluate the performance of \model{} in the RULER benchmark and needle in haystack tasks~\citep{hsieh2024ruler}. The results are reported in \autoref{tab:haystack}. \model{} outperforms all the baselines with +4\% performance gain on average over the second best model, showing higher performance gain of about +6\% in longer sequences. We attribute the superior performance of \model{} to its: (1) more powerful memory management using the hybrid of linear and non-linear recurrence with forget gate, (2) architectural design, in which, we use a two-pass recurrence with \emph{the same memory} to store keys and values. In the results we find two exceptions that \model{} achieves the second best result (i.e., S-NIAH-PK task with 4K and 8K sequence length). These results matches the observation of \citet{yang2024gatedDeltaNet} that simple NIAH task with repeated synthetic context require long-term retention, which a forget gate can damage.

\paragraph{Ablations}
In this section, we perform ablation studies on key components and design choices within the proposed architecture to evaluate the contribution of each to the overall performance. The baseline \model is a 2-pass K-V cache compression layer (presented in equations (\ref{eq:comp-1st-pass}) and (\ref{eq:comp-2nd-pass})) with 125M parameter ($d\text{ = }768$ and $m \text{ = } 64$). Its intermediate activation function: $f(\vx) ~\text{=}~ \op{LN-SiLU}(\vx) ~\text{:=}~ \op{LayerNorm}\left(\op{SiLU}(\vx)\right)$. 
Starting from this baseline configuration, we evaluate the impact of the following modifications (one component at a time):
(1) Removing forget gate,  
(2, 3) Changing the intermediate activation function: normalized $\op{SiLU}$:
$f(\vx) ~\text{=}~ \frac{\op{SiLU}(\vx)}{\|\op{SiLU}(\vx)\|}$ and standard $\op{softmax}$,
(4, 5) Removing the compression layer non-linearity, \textit{i.e.} $\phi(\vx) ~\text{=}~ \vx$ in \eqref{eq:decoding_loss}, which reduces the recurrence in both passes to the Delta Rule. This linear recurrence is then tested with $f~\text{=}~\op{Softmax}$ and $f~\text{=}~\op{LN-SiLU}$.
(6, 7) Varying the memory size,
and 
(8) Using fully non-linear recurrence with mini-batch (chunk) size: $B=1$.
The results, summarized in Table \ref{tab:ablation}, highlight the significance of all these design choices on overall performance, notably, the 2-pass compression, forget gate and the choice of intermediate gate.

\begin{table*}[h!]
\centering
\scriptsize
\caption{Performance comparison of 125M-parameter language models on The Pile (2k context length) and Books (various context lengths) datasets. Baselines include
\method{Transformer}++~\citep{touvron2023llama},
\method{Linear-Attention ~(LA)}~\citep{katharopoulos2020transformers}, \method{DeltaNet}~\citep{yang2024parallelizing}, \method{Gated-DeltaNet}~\citep{yang2024gatedDeltaNet}, \method{Mamba2}~\citep{mamba2}, and \method{TTT}~\citep{sun2024ttt}. 
The best results are \hspace{-1pt} \colorbox{myblue}{highlighted}\hspace{-1pt}.
\label{tab:pile_books}
}
\begin{adjustbox}{width=1.\linewidth}
\setlength{\tabcolsep}{4pt}
\renewcommand{\arraystretch}{1.3}
\begin{tabular}{l|ccccccccc| c}
\toprule
\quad\quad  \textbf{Model}  & \textbf{Pile (2k)} & \textbf{Pile (8k)} &  \textbf{Books (512)} &  \textbf{Books (1k)} & \textbf{Books (2k)} &   \textbf{Books (4k)} & \textbf{Books (8k)} & \textbf{Books (16k)} & \textbf{Books (32k)} & \textbf{Pile (2k)}  \\
 & ppl $\downarrow$  &  ppl $\downarrow$  & ppl $\downarrow$  &  ppl $\downarrow$  & ppl $\downarrow$  &  ppl $\downarrow$  & ppl $\downarrow$  &  ppl $\downarrow$  & ppl $\downarrow$   & ppl $\downarrow$    \\
\midrule
\midrule
\multicolumn{10}{c}{\hspace{-5pt} \textsl{125M params~/~2.4B tokens}}
& 
\multicolumn{1}{c}{\hspace{-5pt} \textsl{350M params~/~7.5B tokens}}
\\
\quad \method{Transformer}++ \quad\quad    &   11.58 &  11.75  & 20.60     & 19.39  & 18.89 & 18.38 & 18.85 & 29.41 & & 8.48\\
\quad \method{Linear-Attention} & 12.43 &  13.26 &  20.84    & 19.96  & 19.60 & 19.64 & 20.34 & 21.87 & 25.07 & 9.16 \\
\quad \method{DeltaNet}        & 11.58 & 11.53 &  20.21     & 19.21  & 18.44 & 17.90 & 18.05 & 18.40 & 19.48 & 8.62 \\
\quad \method{Mamba2}          & 11.47 & 11.35 &  19.96     & 18.89  & 18.19 & 18.07 & 18.23 & 18.40 & 19.33 & 8.56\\
\quad \method{Gated-DeltaNet}  & 11.31 & 10.95 &  19.71     & 18.62  & 17.85 & 17.48 & 17.40 & 17.85 & 18.38 & 8.53 \\
\quad \method{TTT}             & 11.44 & 11.29  & 19.91     & 18.78  & 18.19 & 18.03 & 18.54 & 18.29 & 19.16 & 8.62 \\
\midrule
\quad \method{\model}          & \cellcolor{myblue}{10.87}  & 
\cellcolor{myblue}{10.60} &  
\cellcolor{myblue}{19.09}    & \cellcolor{myblue}{17.92} & \cellcolor{myblue}{17.19} & \cellcolor{myblue}{16.71}  & \cellcolor{myblue}{16.56} & \cellcolor{myblue}{16.58} & \cellcolor{myblue}{17.18}  
& \cellcolor{myblue}{8.26} \\
\bottomrule
\end{tabular}
\end{adjustbox}
\end{table*}

\section{Conclusion}
In this paper, we introduced \model{}, a meta in-context learning framework that learns how to compress the KV cache of attention into a fixed-size memory. \model{} employs a two-pass recurrent memory update process, in which keys and values are stored in a compact memory module. To enhance learning from long context, \model{} further incorporates a gating mechanism that learns how to filter and forget irrelevant past information. Our experimental results across diverse domains—including language modeling, commonsense reasoning, needle-in-a-haystack tasks, and time series data—indicate the superior performance of \model{} compared to modern linear RNNs and Transformers. The results further support the new architectural design of \model, showing that all its components meaningfully contribute to its overall performance. 
By introducing learned, dynamic memory compression, \model offers an effective and efficient solution for long-sequence modeling.

Looking ahead, recent advances in non-linear recurrent models such as Titans~\citep{behrouz2024titans, behrouz2025atlas}, Lattice~\citep{karami2025lattice} offer promising directions for further improving KV cache compression.
Additionally, finetuning pretrained Transformers into RNNs (T2R)~\citep{kasai2021finetuning}  represents another promising direction to explore the capabilities of large pre-trained models with the inference efficiency of recurrent architectures.

\clearpage
\bibliography{References}

@article{karami2025lattice,
  title={Lattice: Learning to Efficiently Compress the Memory},
  author={Karami, Mahdi and Pascanu, Razvan and Mirrokni, Vahab},
  journal={arXiv preprint arXiv:2504.05646},
  year={2025}
}

@inproceedings{karami2024orchid,
  title={Orchid: Flexible and Data-Dependent Convolution for Sequence Modeling},
  author={Karami, Mahdi and Ghodsi, Ali},
    booktitle={Thirty-eighth Conference on Advances in Neural Information Processing Systems},
    url={https://arxiv.org/abs/2402.18508},
  year={2024}
}

@article{karami2019invertible,
  title={Invertible convolutional flow},
  author={Karami, Mahdi and Schuurmans, Dale and Sohl-Dickstein, Jascha and Dinh, Laurent and Duckworth, Daniel},
  journal={Advances in Neural Information Processing Systems},
  volume={32},
  year={2019}
}

@inproceedings{karami2024higen,
  title={{HiGen}: Hierarchical Graph Generative Networks},
  author={Karami, Mahdi},
  booktitle={The Twelfth International Conference on Learning Representations},
  year={2024}
}

@article{karami2017multi,
  title={Multi-view matrix factorization for linear dynamical system estimation},
  author={Karami, Mahdi and White, Martha and Schuurmans, Dale and Szepesv{\'a}ri, Csaba},
  journal={Advances in Neural Information Processing Systems},
  volume={30},
  year={2017}
}

@article{wang2022pretraining,
  title={Pretraining Without Attention},
  author={Wang, Junxiong and Yan, Jing Nathan and Gu, Albert and Rush, Alexander M},
  journal={arXiv preprint arXiv:2212.10544},
  year={2022}
}

@article{fu2023hungry,
  title={Hungry Hungry Hippos: Towards Language Modeling with State Space Models},
  author={Fu, Daniel Y and Dao, Tri and Saab, Khaled K and Thomas, Armin W and Rudra, Atri and R{\'e}, Christopher},
  journal={International Conference on Learning Representations},
  year={2023}
}

@article{poli2023hyena,
  title={Hyena Hierarchy: Towards Larger Convolutional Language Models},
  author={Poli, Michael and Massaroli, Stefano and Nguyen, Eric and Fu, Daniel Y and Dao, Tri and Baccus, Stephen and Bengio, Yoshua and Ermon, Stefano and R{\'e}, Christopher},
  journal={International Conference on Machine Learning},
  year={2023}
}

@inproceedings{chen2021scatterbrain,
  title={Scatterbrain: Unifying Sparse and Low-rank Attention},
  author={Beidi Chen and Tri Dao and Eric Winsor and Zhao Song and Atri Rudra and Christopher R{\'e}},
  booktitle={Advances in Neural Information Processing Systems (NeurIPS)},
  year={2021}
}

@inproceedings{vaswani2018attention,
  title={Attention is all you need},
  author={Vaswani, Ashish and Shazeer, Noam and Parmar, Niki and Uszkoreit, Jakob and Jones, Llion and Gomez, Aidan N and Kaiser, Lukasz and Polosukhin, Illia},
  journal={Advances in neural information processing systems},
  volume={30},
  year={2017}
}

@article{mehta2022long,
  title={Long range language modeling via gated state spaces},
  author={Mehta, Harsh and Gupta, Ankit and Cutkosky, Ashok and Neyshabur, Behnam},
  journal={arXiv preprint arXiv:2206.13947},
  year={2022}
}

@inproceedings{romero2021ckconv,
  title={CKConv: Continuous Kernel Convolution For Sequential Data},
  author={Romero, David W and Kuzina, Anna and Bekkers, Erik J and Tomczak, Jakub Mikolaj and Hoogendoorn, Mark},
  booktitle={International Conference on Learning Representations},
  year={2021}
}

@article{behrouz2024titans,
  title={Titans: Learning to memorize at test time},
  author={Behrouz, Ali and Zhong, Peilin and Mirrokni, Vahab},
  journal={arXiv preprint arXiv:2501.00663},
  year={2024}
}

@inproceedings{kitaevreformer,
  title={Reformer: The Efficient Transformer},
  author={Kitaev, Nikita and Kaiser, Lukasz and Levskaya, Anselm},
  booktitle={International Conference on Learning Representations},
  year={2020},
}

@article{wang2020linformer,
  title={Linformer: Self-attention with linear complexity},
  author={Wang, Sinong and Li, Belinda Z and Khabsa, Madian and Fang, Han and Ma, Hao},
  journal={arXiv preprint arXiv:2006.04768},
  year={2020}
}

@article{choromanski2020rethinking,
  title={Rethinking attention with performers},
  author={Choromanski, Krzysztof and Likhosherstov, Valerii and Dohan, David and Song, Xingyou and Gane, Andreea and Sarlos, Tamas and Hawkins, Peter and Davis, Jared and Mohiuddin, Afroz and Kaiser, Lukasz and others},
  journal={arXiv preprint arXiv:2009.14794},
  year={2020}
}

@article{radford2018improving,
  title={Improving Language Understanding by Generative Pre-Training},
  author={Radford, Alec and Narasimhan, Karthik and Salimans, Tim and Sutskever, Ilya},
  year={2018}
}

@article{beltagy2020longformer,
  title={Longformer: The long-document transformer},
  author={Beltagy, Iz and Peters, Matthew E and Cohan, Arman},
  journal={arXiv preprint arXiv:2004.05150},
  year={2020}
}

@inproceedings{katharopoulos2020transformers,
  title={Transformers are rnns: Fast autoregressive transformers with linear attention},
  author={Katharopoulos, Angelos and Vyas, Apoorv and Pappas, Nikolaos and Fleuret, Fran{\c{c}}ois},
  booktitle={International Conference on Machine Learning},
  pages={5156--5165},
  year={2020},
  organization={PMLR}
}

@article{tay2022efficient,
  title={Efficient transformers: A survey},
  author={Tay, Yi and Dehghani, Mostafa and Bahri, Dara and Metzler, Donald},
  journal={ACM Computing Surveys},
  volume={55},
  number={6},
  pages={1--28},
  year={2022},
  publisher={ACM New York, NY}
}

@article{dosovitskiy2020image,
  title={An image is worth 16x16 words: Transformers for image recognition at scale},
  author={Dosovitskiy, Alexey and Beyer, Lucas and Kolesnikov, Alexander and Weissenborn, Dirk and Zhai, Xiaohua and Unterthiner, Thomas and Dehghani, Mostafa and Minderer, Matthias and Heigold, Georg and Gelly, Sylvain and others},
  journal={arXiv preprint arXiv:2010.11929},
  year={2020}
}

@article{devlin2018bert,
  title={{BERT}: Pre-training of deep bidirectional transformers for language understanding},
  author={Devlin, Jacob and Chang, Ming-Wei and Lee, Kenton and Toutanova, Kristina},
  journal={arXiv preprint arXiv:1810.04805},
  year={2018}
}

@article{dwivedi2020generalization,
  title={A generalization of transformer networks to graphs},
  author={Dwivedi, Vijay Prakash and Bresson, Xavier},
  journal={arXiv preprint arXiv:2012.09699},
  year={2020}
}

@article{child2019generating,
  title={Generating long sequences with sparse transformers},
  author={Child, Rewon and Gray, Scott and Radford, Alec and Sutskever, Ilya},
  journal={arXiv preprint arXiv:1904.10509},
  year={2019}
}

@article{qiu2019blockwise,
  title={Blockwise Self-Attention for Long Document Understanding},
  author={Qiu, Jiezhong and Ma, Hao and Levy, Omer and Yih, Scott Wen-tau and Wang, Sinong and Tang, Jie},
  journal={arXiv preprint arXiv:1911.02972},
  year={2019}
}

@article{tay2020sparse,
  title={Sparse Sinkhorn Attention},
  author={Tay, Yi and Bahri, Dara and Yang, Liu and Metzler, Donald and Juan, Da-Cheng},
  journal={Proceedings of ICML},
  year={2020}
}

@article{roy2020efficient,
  title={Efficient content-based sparse attention with routing transformers},
  author={Roy, Aurko and Saffar, Mohammad and Vaswani, Ashish and Grangier, David},
  journal={Proceedings of TACL},
  year={2020}
}

@article{zaheer2020big,
  title={Big Bird: Transformers for Longer Sequences},
  author={Zaheer, Manzil and Guruganesh, Guru and Dubey, Avinava and Ainslie, Joshua and Alberti, Chris and Ontanon, Santiago and Pham, Philip and Ravula, Anirudh and Wang, Qifan and Yang, Li and others},
  journal={Proceedings of NeurIPS},
  year={2020}
}

@article{parmar2018image,
  title={Image transformer},
  author={Parmar, Niki and Vaswani, Ashish and Uszkoreit, Jakob and Kaiser, {\L}ukasz and Shazeer, Noam and Ku, Alexander and Tran, Dustin},
  journal={Proceedings of ICML 2018},
  year={2018}
}

@article{peng2021random,
  title={Random feature attention},
  author={Peng, Hao and Pappas, Nikolaos and Yogatama, Dani and Schwartz, Roy and Smith, Noah A and Kong, Lingpeng},
  journal={Proceedings of ICLR},
  year={2021}
}

@article{gu2021combining,
  title={Combining recurrent, convolutional, and continuous-time models with linear state space layers},
  author={Gu, Albert and Johnson, Isys and Goel, Karan and Saab, Khaled and Dao, Tri and Rudra, Atri and R{\'e}, Christopher},
  journal={Advances in neural information processing systems},
  volume={34},
  pages={572--585},
  year={2021}
}

@article{arora2024simple,
  title={Simple linear attention language models balance the recall-throughput tradeoff},
  author={Arora, Simran and Eyuboglu, Sabri and Zhang, Michael and Timalsina, Aman and Alberti, Silas and Zinsley, Dylan and Zou, James and Rudra, Atri and R{\'e}, Christopher},
  journal={arXiv preprint arXiv:2402.18668},
  year={2024}
}

@article{gu2023mamba,
  title={Mamba: Linear-time sequence modeling with selective state spaces},
  author={Gu, Albert and Dao, Tri},
  journal={arXiv preprint arXiv:2312.00752},
  year={2023}
}

@article{de2024griffin,
  title={Griffin: Mixing Gated Linear Recurrences with Local Attention for Efficient Language Models},
  author={De, Soham and Smith, Samuel L and Fernando, Anushan and Botev, Aleksandar and Cristian-Muraru, George and Gu, Albert and Haroun, Ruba and Berrada, Leonard and Chen, Yutian and Srinivasan, Srivatsan and others},
  journal={arXiv preprint arXiv:2402.19427},
  year={2024}
}

@inproceedings{
smith2023simplified,
title={Simplified State Space Layers for Sequence Modeling},
author={Jimmy T.H. Smith and Andrew Warrington and Scott Linderman},
booktitle={The Eleventh International Conference on Learning Representations },
year={2023},
url={https://openreview.net/forum?id=Ai8Hw3AXqks}
}

@article{gu2020hippo,
  title={{HiPPO}: Recurrent memory with optimal polynomial projections},
  author={Gu, Albert and Dao, Tri and Ermon, Stefano and Rudra, Atri and R{\'e}, Christopher},
  journal={Advances in neural information processing systems},
  volume={33},
  pages={1474--1487},
  year={2020}
}

@article{beck2024xlstm,
  title={{xLSTM}: Extended Long Short-Term Memory},
  author={Beck, Maximilian and P{\"o}ppel, Korbinian and Spanring, Markus and Auer, Andreas and Prudnikova, Oleksandra and Kopp, Michael and Klambauer, G{\"u}nter and Brandstetter, Johannes and Hochreiter, Sepp},
  journal={arXiv preprint arXiv:2405.04517},
  year={2024}
}

@inproceedings{orvieto2023LRU,
  title={Resurrecting recurrent neural networks for long sequences},
  author={Orvieto, Antonio and Smith, Samuel L and Gu, Albert and Fernando, Anushan and Gulcehre, Caglar and Pascanu, Razvan and De, Soham},
  booktitle={International Conference on Machine Learning},
  pages={26670--26698},
  year={2023},
  organization={PMLR}
}

@inproceedings{mamba2,
  title={Transformers are {SSM}s: Generalized Models and Efficient Algorithms Through Structured State Space Duality},
  author={Dao, Tri and Gu, Albert},
  booktitle={International Conference on Machine Learning (ICML)},
  year={2024}
}

@article{zhang2024gated,
  title={Gated slot attention for efficient linear-time sequence modeling},
  author={Zhang, Yu and Yang, Songlin and Zhu, Ruijie and Zhang, Yue and Cui, Leyang and Wang, Yiqiao and Wang, Bolun and Shi, Freda and Wang, Bailin and Bi, Wei and others},
  journal={arXiv preprint arXiv:2409.07146},
  year={2024}
}

@article{touvron2023llama,
  title={Llama: Open and efficient foundation language models},
  author={Touvron, Hugo and Lavril, Thibaut and Izacard, Gautier and Martinet, Xavier and Lachaux, Marie-Anne and Lacroix, Timoth{\'e}e and Rozi{\`e}re, Baptiste and Goyal, Naman and Hambro, Eric and Azhar, Faisal and others},
  journal={arXiv preprint arXiv:2302.13971},
  year={2023}
}

@article{peng2021abc,
  title={{ABC}: Attention with bounded-memory control},
  author={Peng, Hao and Kasai, Jungo and Pappas, Nikolaos and Yogatama, Dani and Wu, Zhaofeng and Kong, Lingpeng and Schwartz, Roy and Smith, Noah A},
  journal={arXiv preprint arXiv:2110.02488},
  year={2021}
}

@inproceedings{
yang2024parallelizing,
title={Parallelizing Linear Transformers with the Delta Rule over Sequence Length},
author={Songlin Yang and Bailin Wang and Yu Zhang and Yikang Shen and Yoon Kim},
booktitle={The Thirty-eighth Annual Conference on Neural Information Processing Systems},
year={2024},
url={https://openreview.net/forum?id=y8Rm4VNRPH}
}

@inproceedings{
yang2024gatedattn,
title={Gated Linear Attention Transformers with Hardware-Efficient Training},
author={Songlin Yang and Bailin Wang and Yikang Shen and Rameswar Panda and Yoon Kim},
booktitle={Forty-first International Conference on Machine Learning},
year={2024},
url={https://openreview.net/forum?id=ia5XvxFUJT}
}

@article{yang2024gatedDeltaNet,
  title={Gated Delta Networks: Improving Mamba2 with Delta Rule},
  author={Yang, Songlin and Kautz, Jan and Hatamizadeh, Ali},
  journal={arXiv preprint arXiv:2412.06464},
  year={2024}
}

@inproceedings{
kacham2024polysketchformer,
title={PolySketchFormer: Fast Transformers via Sketching Polynomial Kernels},
author={Praneeth Kacham and Vahab Mirrokni and Peilin Zhong},
booktitle={Forty-first International Conference on Machine Learning},
year={2024},
url={https://openreview.net/forum?id=ghYrfdJfjK}
}

@inproceedings{hua2022transformer,
  title={Transformer quality in linear time},
  author={Hua, Weizhe and Dai, Zihang and Liu, Hanxiao and Le, Quoc},
  booktitle={International conference on machine learning},
  pages={9099--9117},
  year={2022},
  organization={PMLR}
}

@article{sun2024ttt,
  title={Learning to (learn at test time): {RNNs} with expressive hidden states},
  author={Sun, Yu and Li, Xinhao and Dalal, Karan and Xu, Jiarui and Vikram, Arjun and Zhang, Genghan and Dubois, Yann and Chen, Xinlei and Wang, Xiaolong and Koyejo, Sanmi and others},
  journal={arXiv preprint arXiv:2407.04620},
  year={2024}
}

@inproceedings{schlag2021linear,
  title={Linear transformers are secretly fast weight programmers},
  author={Schlag, Imanol and Irie, Kazuki and Schmidhuber, J{\"u}rgen},
  booktitle={International Conference on Machine Learning},
  pages={9355--9366},
  year={2021},
  organization={PMLR}
}

@article{andrychowicz2016learning,
  title={Learning to learn by gradient descent by gradient descent},
  author={Andrychowicz, Marcin and Denil, Misha and Gomez, Sergio and Hoffman, Matthew W and Pfau, David and Schaul, Tom and Shillingford, Brendan and De Freitas, Nando},
  journal={Advances in neural information processing systems},
  volume={29},
  year={2016}
}

@article{schmidhuber1992learning,
  title={Learning to control fast-weight memories: An alternative to recurrent nets},
  author={Schmidhuber, JH},
  journal={Neural Computation},
  year={1992}
}

@article{gonzalez2024towards,
  title={Towards Scalable and Stable Parallelization of Nonlinear RNNs},
  author={Gonzalez, Xavier and Warrington, Andrew and Smith, Jimmy TH and Linderman, Scott W},
  journal={arXiv preprint arXiv:2407.19115},
  year={2024}
}

@article{lim2023parallelizing,
  title={Parallelizing non-linear sequential models over the sequence length},
  author={Lim, Yi Heng and Zhu, Qi and Selfridge, Joshua and Kasim, Muhammad Firmansyah},
  journal={arXiv preprint arXiv:2309.12252},
  year={2023}
}

@article{blelloch1990prefix,
  title={Prefix sums and their applications},
  author={Blelloch, Guy E},
  year={1990},
  publisher={School of Computer Science, Carnegie Mellon University Pittsburgh, PA, USA}
}

@article{yun2019graph,
  title={Graph transformer networks},
  author={Yun, Seongjun and Jeong, Minbyul and Kim, Raehyun and Kang, Jaewoo and Kim, Hyunwoo J},
  journal={Advances in neural information processing systems},
  volume={32},
  year={2019}
}

@article{gao2020pile,
  title={The pile: An 800gb dataset of diverse text for language modeling},
  author={Gao, Leo and Biderman, Stella and Black, Sid and Golding, Laurence and Hoppe, Travis and Foster, Charles and Phang, Jason and He, Horace and Thite, Anish and Nabeshima, Noa and others},
  journal={arXiv preprint arXiv:2101.00027},
  year={2020}
}

@inproceedings{arnab2021vivit,
  title={Vivit: A video vision transformer},
  author={Arnab, Anurag and Dehghani, Mostafa and Heigold, Georg and Sun, Chen and Lu{\v{c}}i{\'c}, Mario and Schmid, Cordelia},
  booktitle={Proceedings of the IEEE/CVF international conference on computer vision},
  pages={6836--6846},
  year={2021}
}

@article{wang2025testTimeRegression,
  title={Test-time regression: a unifying framework for designing sequence models with associative memory},
  author={Wang, Ke Alexander and Shi, Jiaxin and Fox, Emily B},
  journal={arXiv preprint arXiv:2501.12352},
  year={2025}
}

@incollection{thrun1998learning,
  title={Learning to learn: Introduction and overview},
  author={Thrun, Sebastian and Pratt, Lorien},
  booktitle={Learning to learn},
  pages={3--17},
  year={1998},
  publisher={Springer}
}

@article{liu2022bome,
  title={Bome! bilevel optimization made easy: A simple first-order approach},
  author={Liu, Bo and Ye, Mao and Wright, Stephen and Stone, Peter and Liu, Qiang},
  journal={Advances in neural information processing systems},
  volume={35},
  pages={17248--17262},
  year={2022}
}

@article{chen2022gradient,
  title={Gradient-based bi-level optimization for deep learning: A survey},
  author={Chen, Can and Chen, Xi and Ma, Chen and Liu, Zixuan and Liu, Xue},
  journal={arXiv preprint arXiv:2207.11719},
  year={2022}
}

@misc{behrouz2025itsconnected,
      title={It's All Connected: A Journey Through Test-Time Memorization, Attentional Bias, Retention, and Online Optimization}, 
      author={Ali Behrouz and Meisam Razaviyayn and Peilin Zhong and Vahab Mirrokni},
      year={2025},
      eprint={2504.13173},
      archivePrefix={arXiv},
      primaryClass={cs.LG},
      url={https://arxiv.org/abs/2504.13173}, 
}

@article{behrouz2025atlas,
  title={Atlas: Learning to optimally memorize the context at test time},
  author={Behrouz, Ali and Li, Zeman and Kacham, Praneeth and Daliri, Majid and Deng, Yuan and Zhong, Peilin and Razaviyayn, Meisam and Mirrokni, Vahab},
  journal={arXiv preprint arXiv:2505.23735},
  year={2025}
}

@inproceedings{
hsieh2024ruler,
title={{RULER}: What{\textquoteright}s the Real Context Size of Your Long-Context Language Models?},
author={Cheng-Ping Hsieh and Simeng Sun and Samuel Kriman and Shantanu Acharya and Dima Rekesh and Fei Jia and Boris Ginsburg},
booktitle={First Conference on Language Modeling},
year={2024},
url={https://openreview.net/forum?id=kIoBbc76Sy}
}

@article{singhania2024loki,
  title={Loki: Low-rank keys for efficient sparse attention},
  author={Singhania, Prajwal and Singh, Siddharth and He, Shwai and Feizi, Soheil and Bhatele, Abhinav},
  journal={arXiv preprint arXiv:2406.02542},
  year={2024}
}

@book{haykin2002adaptive,
  title={Adaptive filter theory},
  author={Haykin, Simon S},
  year={2002},
  publisher={Pearson Education India}
}

@book{bengio1990learning,
  title={Learning a synaptic learning rule},
  author={Bengio, Yoshua and Bengio, Samy and Cloutier, Jocelyn},
  year={1990},
  publisher={Citeseer}
}

@incollection{widrow1988adaptive,
  title={Adaptive switching circuits},
  author={Widrow, Bernard and Hoff, Marcian E},
  booktitle={Neurocomputing: foundations of research},
  pages={123--134},
  year={1988}
}

@article{raffel2020exploring,
  title={Exploring the limits of transfer learning with a unified text-to-text transformer},
  author={Raffel, Colin and Shazeer, Noam and Roberts, Adam and Lee, Katherine and Narang, Sharan and Matena, Michael and Zhou, Yanqi and Li, Wei and Liu, Peter J},
  journal={Journal of machine learning research},
  volume={21},
  number={140},
  pages={1--67},
  year={2020}
}

@article{li2022makes,
  title={What makes convolutional models great on long sequence modeling?},
  author={Li, Yuhong and Cai, Tianle and Zhang, Yi and Chen, Deming and Dey, Debadeepta},
  journal={arXiv preprint arXiv:2210.09298},
  year={2022}
}

@article{kasai2021finetuning,
  title={Finetuning pretrained transformers into rnns},
  author={Kasai, Jungo and Peng, Hao and Zhang, Yizhe and Yogatama, Dani and Ilharco, Gabriel and Pappas, Nikolaos and Mao, Yi and Chen, Weizhu and Smith, Noah A},
  journal={arXiv preprint arXiv:2103.13076},
  year={2021}
}

@article{sun2024you,
  title={You only cache once: Decoder-decoder architectures for language models},
  author={Sun, Yutao and Dong, Li and Zhu, Yi and Huang, Shaohan and Wang, Wenhui and Ma, Shuming and Zhang, Quanlu and Wang, Jianyong and Wei, Furu},
  journal={Advances in Neural Information Processing Systems},
  volume={37},
  pages={7339--7361},
  year={2024}
}

@inproceedings{karami2025MSSSM,
  title={MS-SSM: A Multi-Scale State Space Model for Efficient Sequence Modeling},
  author={Karami, Mahdi and Behrouz, Ali and Zhong, Peilin and Pascanu, Razvan and Mirrokni, Vahab},
  booktitle={Second Conference on Language Modeling},
  year={2025}
}
\bibliographystyle{colm2025_conference}

\clearpage

\appendix

\section{Experiment Details} \label{apdx:arch-detail}

\head{Architectural Details}
We provide the architectural details of our model in \autoref{tab:exp-details}.

\begin{table*}[h!]
    \centering
    \caption{Architectural Details.}
    \label{tab:exp-details}
    \resizebox{0.5\linewidth}{!}{
    \begin{tabular}{c c c c c c}
    \toprule
         Model    & Block & Dim & Head & Peak LR & Token\\
         \midrule
         \midrule
        125M & 12 & 768 & 12 & 3e-3 & 2.4B\\
        170M & 12 & 768 & 16 & 3e-3 & 15B\\
        350M & 24 & 1024 & 16 & 1.5e-3 & 15B\\
        780M & 24 & 1536 & 16 & 1.25e-3 & 30B\\
    \toprule
    \end{tabular}
    }
\end{table*}

\head{The Choice of Datasets and Training} We train our models on three different datasets. Our goal is to highlight the generalizability and power of Trellis across diverse datasets. For smaller and medium-sized models (125M and 350M parameters), we included a broader set of baselines and conducted detailed ablation studies. Due to computational constraints, we limited the training of larger models (780M and 1B parameters) to comparisons with the most relevant baselines. Note that Tables 1 and 3 present results for models with 780M and 1B parameters, respectively, to show Trellis’s performance at larger scales. The results in the bottom row of Figure 2 are based on models trained on The Pile and Books datasets. All ablation studies and Table 4 are also conducted using models trained on The Pile. The models used in the remaining experiments (including Table 3 and Figure 3) are trained on the C4 dataset.

\end{document}